# Convex-constrained Sparse Additive Modeling and Its Extensions

Junming Yin* and Yaoliang Yu†

May 1, 2017

**Abstract**

Sparse additive modeling is a class of effective methods for performing high-dimensional non-parametric regression. In this work we show how shape constraints such as convexity/concavity and their extensions, can be integrated into additive models. The proposed sparse difference of convex additive models (SDCAM) can estimate most continuous functions without any *a priori* smoothness assumption. Motivated by a characterization of difference of convex functions, our method incorporates a natural regularization functional to avoid overfitting and to reduce model complexity. Computationally, we develop an efficient backfitting algorithm with linear per-iteration complexity. Experiments on both synthetic and real data verify that our method is competitive against state-of-the-art sparse additive models, with improved performance in most scenarios.

## 1 Introduction

Regression, which aims at predicting a real-valued response from a set of covariates, is an important form of supervised learning and has been widely studied in machine learning and statistics. More precisely, consider the following model:

$$Y = m(X) + \xi,$$

where the covariate $X \in \mathbb{R}^p$ and the noise $\xi \in \mathbb{R}$ are both random. It is common in practice to assume $\mathbb{E}(\xi|X) = 0$, hence $m(\mathbf{x}) = \mathbb{E}(Y|X = \mathbf{x})$ is the regression function that we aim to estimate based on a set of data samples $\{(\mathbf{x}_i, y_i) : i = 1, \ldots, n\}$. Asymptotic consistency for certain estimators (such as least-squares) is well-known, although the rate of convergence can be arbitrarily slow [GKKW02]. Meaningful and precise rates of convergence have been obtained if the regression function $m$ is smooth, or, as we see more recently [GJ14], if $m$ satisfies certain shape restrictions (e.g., linear, monotonic, convex, or log-concave). However, even under such extra regularity conditions, the rate of convergence, quite disappointingly, still degrades quickly as the dimension $p$ increases—a phenomenon widely known as the "curse of dimensionality."

Additive modeling [HT90] is a class of convenient tools to combat the curse of dimensionality, by making further assumption that the regression function $m$ can be decomposed (approximately) into a sum of *univariate* component functions, each acting on a separate covariate. More precisely,

$$m(\mathbf{x}) \approx \mu + \sum_{j=1}^{p} f_j(x_j), \tag{1}$$

---

*Eller College of Management, University of Arizona, Tucson, AZ 85721. Email: junmingy@email.arizona.edu
†School of Computer Science, University of Waterloo, Waterloo, ON, N2L3G1. Email: yaoliang.yu@uwaterloo.ca

where $\mu \in \mathbb{R}$ is the intercept and each component $f_j$ is a univariate function of $x_j$ only. By restricting itself to the additive form (1), the class of additive models can bring in several advantages:

- dimensionality: a celebrated result of [Sto85] reveals that when each component $f_j$ is smooth, the additive models can achieve the same (optimal) rate of convergence for general $p$ as they do for $p = 1$.

- interpretability: each component function can be easily visualized and inspected.

- computability: by decomposing the multivariate estimation problem into a set of univariate ones we can greatly simplify the underlying computations.

Previous works have also studied additive models where some interactions among covariates are allowed [HT90, TKGK16, KY16].

Our main interest in this work is to incorporate certain shape constraints of component functions, such as convexity and its extension, into the procedure of estimating high dimensional additive models. In modern applications it is not uncommon that the number of covariates $p$ is comparable with or even greatly exceeds the number of samples $n$ that one can possibly (afford to) collect. However, the true regression function may only depend on a small number of important covariates, at least approximately. In the case of additive modeling, it amounts to identifying those few nonzero functional components in (1) based on a set of data samples. It is also equally important that the regression function is usually subject to certain *a priori* regularity conditions. For instance, many functions that arise in econometrics [Var82, Var84], actuarial science [Gre56], statistics [GJ14], and optimization [HD13, HPD14] *tend* to be convex/concave or monotonic [BBBB72]. However, such *a priori* knowledge can be imprecise and sometimes even wrong, e.g., convex functions are mistakenly believed to be concave. Therefore, we need our estimation procedure of additive models to be robust against the prior knowledge of component functions, such as their shape constraints. More precisely, we make the following contributions in this work:

- We propose a sparse convex additive model to estimate convex (and monotonic) component functions in high dimensional additive modeling.

- To address potential robustness issue, we next propose the sparse difference of convex additive models (SDCAM) that can estimate most continuous functions, being smooth or not, that we use in practice. We discover three surprising properties of SDCAM: (1) If naively formulated, SDCAM is bound to overfit the training data; (2) By relating to a less-known characterization of difference of convex functions, we show how to add a very natural regularization functional to SDCAM to avoid overfitting and to reduce model complexity; (3) Despite the well-known fact that minimizing a difference of convex function is generally intractable, estimating a difference of convex function, as in SDCAM, can be done very efficiently.

- We propose an efficient backfitting algorithm for SDCAM with linear per-iteration complexity. It includes many prior algorithms as special cases.

- We test SDCAM on both synthetic and real data. Overall, SDCAM achieved competitive model selection and prediction accuracy against state-of-the-art sparse additive models. SDCAM also leads to some interesting new discovery on the Boston housing data.

This paper proceeds as follows. In the next section §2, we first introduce some background and related work. Then, we present the new sparse difference of convex additive models (SDCAM) in §3. Experimental results are reported in §4. Finally, we conclude in §5.

## 2 Background and Related Work

In this section, we briefly review shape-constrained functions that are used in additive modeling before. Recall that we are interested in estimating a multivariate function[1] $m : [0,1]^p \to \mathbb{R}$, based on *i.i.d.* samples $\{(\mathbf{x}_i, y_i) : i = 1, \ldots, n\}$ from the following model

$$Y = m(X) + \xi,$$

where $\xi$ is a random noise that is independent of $X$. To combat the curse of dimensionality, additive models try to find an additive approximation of the regression function

$$m(\mathbf{x}) = \mathbb{E}(Y|X = \mathbf{x}) \approx \mu + \sum_{j=1}^{p} f_j(x_j).$$

For the sake of model identifiability, we assume each univariate component function $f_j$ is centered, i.e.,

$$\mathbb{E}(f_j(X_j)) = 0, j = 1, \ldots, p.$$

As a consequence, an unbiased estimator of $\mu$ is simply the sample mean of the observed responses $\{y_i\}$. In what follows, we always first subtract this sample mean from the responses $y_i$ hence the intercept $\mu$ can be omitted from the model.

### 2.1 Isotonic Functions

[Bac89] first studied the isotonic additive modeling, in which each component function $f_j$ is restricted to be isotonic, i.e., in the function class

$$\mathcal{M}_j := \{f : [0,1] \to \mathbb{R} \,|\, \mathbb{E}(f(X_j)) = 0, f \text{ is increasing}\}.$$

Given $n$ samples $\{(\mathbf{x}_i, y_i) : i = 1, \ldots, n\} \subseteq [0,1]^p \times \mathbb{R}$, the isotonic additive regression solves the following least squares estimation problem:

$$\min_{\forall j, f_j \in \mathcal{M}_j} \sum_{i=1}^{n} \left( y_i - \sum_{j=1}^{p} f_j(x_{ij}) \right)^2. \qquad (2)$$

Since the objective function depends on each $f_j$ only at observed values, we can simplify (2) as follows: First, we sort each of the $p$ covariates $\mathbf{x}_j$ separately in increasing order. Throughout this work we use the **tilde notation** $\tilde{\mathbf{x}}_j$ to represent this operation (w.r.t. $\mathbf{x}_j$), that is, $\tilde{\mathbf{x}}_j = \text{sort}(\mathbf{x}_j)$ such that

$$\tilde{x}_{1j} \leq \tilde{x}_{2j} \leq \cdots \leq \tilde{x}_{nj}.$$

Note that different covariates may require different permutations of the indexes in order to be sorted increasingly. Next, we introduce the component fits $\mathbf{z}_j \in \mathbb{R}^n$ on the observed values: $z_{ij} = f_j(x_{ij}), i = 1, \ldots, n$, so that the monotonicity of the function $f_j$ reduces to that of the vector $\mathbf{z}_j$. Lastly, we approximate the centering constraint $\mathbb{E}(f_j(X_j)) = 0$ by the empirical average over the samples. Thus, we arrive at the following *finite-dimensional* least squares problem:

$$\min_{\forall j, \tilde{\mathbf{z}}_j \in \mathcal{I} \cap \mathcal{H}} \sum_{i=1}^{n} \left( y_i - \sum_{j=1}^{p} z_{ij} \right)^2, \tag{3}$$

where $\mathcal{H} := \{\mathbf{z} \in \mathbb{R}^n : \sum_{i=1}^n z_i = 0\}$ is the centering constraint and

$$\mathcal{I} := \{\mathbf{z} \in \mathbb{R}^n : z_1 \leq z_2 \leq \cdots \leq z_n\} \tag{4}$$

is the isotonic (monotonic) cone. Note that all shape constraints in this work are with respect to the covariates $\mathbf{x}_j$, hence we use the tilde notation $\tilde{\mathbf{z}}_j$ to indicate this dependence. That is, $\tilde{z}_{ij} = f_j(\tilde{x}_{ij})$ is a permuted version of $\mathbf{z}_j$, according to $\mathbf{x}_j$. In what follows, one should keep in mind that $\tilde{\mathbf{z}}_j = P_j \mathbf{z}_j$ where the permutation matrix $P_j$ satisfies $\tilde{\mathbf{x}}_j = P_j \mathbf{x}_j$.

---
[1] For simplicity, we assume the domain of the regression function is $[0,1]^p$ throughout the paper.

[Bac89] developed a backfitting (i.e., block coordinate descent) algorithm for solving (3) while ignoring the centering constraint $\mathbf{z}_j \in \mathcal{H}$: in each step, all components except for one $\mathbf{z}_j$ are fixed, and the well-known pool-adjacent-violators (PAV) algorithm [BBBB72] is applied to optimize $\mathbf{z}_j$ under the isotonic constraint $\mathcal{I}$. Oracle property of the backfitting estimator was later investigated by [MY07].

Similar to standard additive models, isotonic additive modeling can fail in the high-dimensional setting ($p > n$). [FM12] proposed the LASSO-Isotone (LISO) estimator, with the aim to incorporate the idea of sparsity into isotonic additive regression. Specifically, the total variation of component fits $\mathbf{z}_j$ were added to the objective function of problem (3) as a LASSO-style penalty:

$$\min_{\forall j, \tilde{\mathbf{z}}_j \in \mathcal{I} \cap \mathcal{H}} \sum_{i=1}^{n} \left( y_i - \sum_{j=1}^{p} z_{ij} \right)^2 + \sum_{j=1}^{p} \lambda_t \|\tilde{\mathbf{z}}_j\|_{\mathsf{tv}}, \tag{5}$$

where the total variation

$$\|\tilde{\mathbf{z}}_j\|_{\mathsf{tv}} := \sum_{i=2}^{n} |\tilde{z}_{ij} - \tilde{z}_{i-1,j}|$$

simplifies to $\tilde{z}_{nj} - \tilde{z}_{1j}$ under the isotonic constraint $\tilde{\mathbf{z}}_j \in \mathcal{I}$. An iterative procedure called LISO-backfitting was applied to solve the LISO optimization problem (5).

## 2.2 Functions of Bounded Variation

The set of all isotonic functions $\mathcal{M}_j$ forms a convex cone in the vector space of all functions

$$\mathcal{F} := \{f \mid f : [0,1] \to \mathbb{R}\}.$$

It is thus natural to consider the subspace generated by isotonic functions, i.e., all functions that can be written as the difference of two increasing functions:

$$\mathcal{BV}_j := \{f \in \mathcal{F} \mid f = f_1 - f_2, f_1 \in \mathcal{M}_j, f_2 \in \mathcal{M}_j\}.$$

As is well-known in real analysis [Apo74, Theorem 6.13], the above class of functions coincides with the class of functions of bounded variation, i.e., functions $f$ whose total variation is finite:

$$\|f\|_{\mathsf{tv}} := \sup_{\mathcal{P}} \sum_{i=2}^{n} |f(x_i) - f(x_{i-1})| < \infty,$$

where the supremum is taken over $n$ and all partitions $\mathcal{P} = \{0 \leq x_1 \leq x_2 \leq \cdots \leq x_n \leq 1\}$ of $[0,1]$.

The class of bounded variation functions motivates the development of the fused lasso additive model (FLAM), in which each component function $f_j$ is estimated to be piecewise constant with a small number of adaptively chosen knots [PWS16]. FLAM solves the following regularized least squares problem using block coordinate descent:

$$\min_{\forall j, \mathbf{z}_j \in \mathcal{H}} \sum_{i=1}^{n} \left(y_i - \sum_{j=1}^{p} z_{ij}\right)^2 + \sum_{j=1}^{p} (\lambda_t \|\tilde{\mathbf{z}}_j\|_{\mathsf{tv}} + \lambda_s \|\mathbf{z}_j\|_2),$$

where the total variation seminorm $\|\tilde{\mathbf{z}}_j\|_{\mathsf{tv}}$ encourages each component function to be piecewise constant and the $\ell_2$ norm $\|\mathbf{z}_j\|_2$ imposes sparsity on the component functions to conduct variable selection. The centering constraint $\tilde{\mathbf{z}}_j \in \mathcal{H}$, however, is dropped in the implementation of FLAM, and the intercept $\mu$ has to be re-estimated in each iteration.

## 3 Convex-constrained Sparse Additive Models

Motivated by the previous work, in this section we study a related but different shape constraint, namely convexity, in sparse additive modeling. That is, each non-zero component function $f_j$ is restricted to the following function class

$$\mathcal{C}_j := \{f : [0,1] \to \mathbb{R} \mid \mathbb{E}(f(X_j)) = 0, f \text{ is convex}\}.$$

Although there is a certain overlap between the two function classes $\mathcal{M}_j$ and $\mathcal{C}_j$, their "sizes" are quite different: the metric entropy of $\mathcal{M}_j$ (under the $\ell_\infty$ distance) is known to be of the order $\epsilon^{-1}$[vdVW00, Theorem 2.7.5], while the metric entropy of the class of convex and uniformly bounded functions is of the significantly smaller order $\epsilon^{-1/2}$ [GS13].

[CS16] studied a related problem where each component function in a generalized additive (index) model is required to obey a certain shape restriction, such as monotonicity, convexity, or the combination of the two (see their Table 1). In the case that it is a standard additive model and all the component functions are constrained to be convex, the resulting convex additive regression solves the following constrained least squares problem:

$$\min_{\forall j, f_j \in \mathcal{C}_j} \sum_{i=1}^{n} \left(y_i - \sum_{j=1}^{p} f_j(x_{ij})\right)^2. \tag{6}$$

[CS16] applied an active set algorithm to solve (6) using basis expansion, which results in $np$ parameters hence unfortunately can fail to work well in the high-dimensional setting. In fact, in their simulation study and real data examples, the dimensionality of covariates being considered is no larger than 10.

## 3.1 Sparse Convex Additive Model (SCAM)

To deal with high dimensionality under the reasonable assumption that most covariates are irrelevant, we propose the following penalized optimization problem:

$$\min_{\forall j, f_j \in \mathcal{C}_j} \sum_{i=1}^{n} \left(y_i - \sum_{j=1}^{p} f_j(x_{ij})\right)^2 + \lambda_s \sum_{j=1}^{p} \|f_j\|_2, \tag{7}$$

where $\|f_j\|_2 := \sqrt{\mathbb{E}(f_j^2(X_j))}$ is the $L_2$ norm of component function $f_j$ w.r.t. the marginal distribution of $X_j$. When $p$ is large and $\lambda_s$ is chosen appropriately, the $L_2$ norm in (7) will set many irrelevant component functions $f_j$ to zero hence achieving model selection. Our model (7) here is inspired by SpAM [RLLW08], which first considered the $L_2$ group norm in nonparametric additive modeling. The difference is that SpAM relies on linear smoothers to estimate *smooth* component functions while we are interested in getting *convex* component functions $\mathcal{C}_j$—a shape constraint that SpAM cannot explicitly take into account. We note in passing that a similar model to (7) (with group $\ell_\infty$ norm instead) appeared as the first stage of the screening procedure in the recent work of [XCL16],

Following the discussion from §2.1, we can simplify the infinite-dimensional problem (7) by first sorting each of the $p$ covariates[2]. Again, we use the tilde notation to signal this sorting operation w.r.t. the corresponding covariate. Then, we introduce the fits $\mathbf{z}_j \in \mathbb{R}^n$ on the observed values, i.e., $z_{ij} = f_j(x_{ij}), i = 1\ldots,n$, and make the substitution in the objective of (7). Lastly, we approximate the centering constraint $\mathbb{E}(f_j(X_j)) = 0$ and $L_2$ norm $\|f_j\|_2 := \sqrt{\mathbb{E}(f_j^2(X_j))}$ by the empirical average over the samples, leading to the *finite-dimensional* problem of the sparse convex additive model (SCAM):

$$\min_{\forall j, \tilde{\mathbf{z}}_j \in \mathcal{K}_j \cap \mathcal{H}} \sum_{i=1}^{n} \left(y_i - \sum_{j=1}^{p} z_{ij}\right)^2 + \lambda_s \sum_{j=1}^{p} \|\mathbf{z}_j\|_2, \tag{8}$$

where again $\mathcal{H} := \{\mathbf{z} \in \mathbb{R}^n : \sum_{i=1}^{n} z_i = 0\}$ is the empirical centering constraint and

$$\mathcal{K}_j := \left\{\mathbf{z} \in \mathbb{R}^n : \frac{z_2 - z_1}{\tilde{x}_{2j} - \tilde{x}_{1j}} \leq \cdots \leq \frac{z_n - z_{n-1}}{\tilde{x}_{nj} - \tilde{x}_{n-1,j}}\right\}. \tag{9}$$

Here, we use an important observation from the celebrated work of [Hil54], namely, the convex cone constraint $\mathcal{K}_j$ is both sufficient and necessary to guarantee the underlying component function $f_j$ to be convex. Note that had we not assumed the additive form $m(\mathbf{x}) \approx \sum_j f_j(x_j)$, we would have to deal with $n^2$ linear constraints (see for instance [BV04, §6.5.5]) instead of only $n$ simple ones. Therefore, the additive modeling assumption is not only useful in combating the curse of dimensionality; it also leads to computationally more convenient formulations.

Moreover, we can further combine monotonicity with convexity, as studied in [CS16, Table 1]. For instance, if some component function $f_j$ is believed to be both convex and monotonically increasing, then we need only replace the cone $\mathcal{K}_j$ defined in (9) by

$$\mathcal{K}'_j := \left\{\mathbf{z} \in \mathbb{R}^n : 0 \leq \frac{z_2 - z_1}{\tilde{x}_{2j} - \tilde{x}_{1j}} \leq \cdots \leq \frac{z_n - z_{n-1}}{\tilde{x}_{nj} - \tilde{x}_{n-1,j}}\right\}. \tag{10}$$

---

[2]For simplicity, we assume that the observed values of each covariate are distinct. It is straightforward to extend to the case where ties can possibly arise.

As we shall see, this change will have very minimal effect on our backfitting algorithm hence below we do not discuss such straightforward extensions in details.

## 3.2 Sparse Difference of Convex Additive Model (SDCAM)

The qualitative convex assumption in (7) is natural in many applications, but perhaps we can only expect it to hold *approximately*? Besides, what if we were wrong and the component function $f_j$ is in fact *concave*? In this section we propose a convenient extension to address such robustness issue.

The idea is to work with a larger class of functions. In particular, in parallel with §2.2, we consider class of difference of convex (DC) functions:

$$\mathcal{DC}_j := \{f \in \mathcal{F} \mid f = f_1 - f_2, f_1 \in \mathcal{C}_j, f_2 \in \mathcal{C}_j\}.$$

Clearly, $\mathcal{DC}_j$ is the vector space generated by $\mathcal{C}_j$. From the definition it is also clear that all convex functions, as well as all concave functions, are DC. In fact, most continuous functions (convex or not, smooth or not) we use in practice are DC. For comparison purpose we state the following fact:

**Theorem 1.** $\mathcal{M}_j \subset \mathcal{BV}_j \supset \mathcal{DC}_j \supset \mathcal{C}_j$.

As an example, we note that simple step functions are of bounded variation but they are not DC, since all DC functions are continuous (except perhaps at the boundaries of the domain). The difference is large in terms of metric entropy: $O(\epsilon^{-1})$ for $\mathcal{BV}_j$ versus $O(\epsilon^{-1/2})$ for $\mathcal{DC}_j$. As a result, searching in the space of DC functions is "easier" than in the space of bounded variation functions.

However, if we naively replace $\mathcal{C}_j$ with $\mathcal{DC}_j$ in the constraint of (7), the solution will severely overfit to the data, due to the following result:

**Theorem 2.** *For any sample $\{(\mathbf{x}_i, y_i) : i = 1, \ldots, n\} \subseteq [0,1]^p \times \mathbb{R}$ such that $\mathbf{x}_i \neq \mathbf{x}_j$ for all $1 \leq i \neq j \leq n$, there always exists a multivariate[3] DC function $f : [0,1]^p \to \mathbb{R}$ such that for all $i = 1, \ldots, n$, $f(\mathbf{x}_i) = y_i$.*

*Proof.* We only need to consider the special case of $p = 1$, which can be proved much more easily. However, we provide a proof for any $p$ because this will provide another justification for the use of additive models.

Since $\mathbf{x}_i \neq \mathbf{x}_j$ for any $i \neq j$, we can put a bump function $f_i$ around a small neighborhood $N_i$ of each $\mathbf{x}_i$, such that $N_i \cap N_j = \emptyset$ for any $i \neq j$ and $f_i(\mathbf{z}) = 0$ iff $\mathbf{z} \notin N_i$. For instance, we can choose

$$f_i(\mathbf{z}) = \begin{cases} y_i e^{\frac{p}{\delta^2}} \prod_{j=1}^p e^{-\frac{1}{\delta^2 - (z_j - x_{ij})^2}}, & \text{if } \|\mathbf{z} - \mathbf{x}_i\|_\infty < \delta \\ 0, & \text{otherwise} \end{cases}$$

for some sufficiently small positive $\delta$. Note that $f_i(\mathbf{x}_i) = y_i$ and $f_i$ is twice (in fact infinitely many times) continuously differentiable. Now consider $f(\mathbf{z}) = \sum_{i=1}^n f_i(\mathbf{z})$. Since the functions $f_i$ have non-overlapping support, $f(\mathbf{x}_i) = f_i(\mathbf{x}_i) = y_i$ for all $i = 1, \ldots, n$. Moreover, the Hessian $\nabla^2 f$ is bounded from below. Therefore, for $\gamma$ sufficiently large, the function $f(\mathbf{z}) + \gamma \|\mathbf{z}\|_2^2$ is convex, which implies that $f$ is DC.

The centering constraint $\mathbb{E}(f(X_j)) = 0$ can also be satisfied, by choosing a centered bump function, or we can add an extra point $\mathbf{x}_0 \notin \{\mathbf{x}_1, \ldots, \mathbf{x}_n\}$ and an additional component $f_0$ with suitable $y_0$ to cancel the integrals of $f_1, \ldots, f_n$. □

---

[3] The definition of multivariate DC functions is a straightforward extension of the univariate case.

Therefore, naively estimating DC functions as in (7) would always lead to zero training error, effectively memorizing the training samples. The fundamental reason is that the class of DC functions is too rich. Thus, we need a complexity measure to penalize those estimators that use very "complex" DC functions. Fortunately, this can be achieved through the following characterization of DC functions of one variable:

**Theorem 3** ([RV73, §14]). *Let $f : [0,1] \to \mathbb{R}$ be a function with finite one-sided derivatives at 0 and 1. Then $f$ is DC iff*

$$\|f\|_{\mathcal{DC}} := \sup_{\mathcal{P}} \sum_{i=2}^{n-1} \left| \frac{f(x_{i+1}) - f(x_i)}{x_{i+1} - x_i} - \frac{f(x_i) - f(x_{i-1})}{x_i - x_{i-1}} \right| < \infty,$$

*where the supremum is taken over $n$ and the partitions $\mathcal{P} = \{0 \leq x_1 < x_2 < \cdots < x_n \leq 1\}$ of $[0,1]$.*

In particular, if $f$ is convex, then $\|f\|_{\mathcal{DC}} = f'_-(1) - f'_+(0)$, which explains the assumption of finite one-sided derivatives. Moreover, $\|f\|_{\mathcal{DC}} = 0$ iff $f$ is affine, i.e., $f(x) = ax + b$ for some constant $a$ and $b$. Essentially, the seminorm $\|f\|_{\mathcal{DC}}$ measures how fast the slope of $f$ changes. If $f$ is twice continuously differentiable, then

$$\|f\|_{\mathcal{DC}} = \int_0^1 |f''(x)| \, \mathrm{d}x.$$

Note that it is crucial here that $f$ is a *univariate* function. For multivariate DC functions, we still do not have a convenient characterization as in Theorem 3.

Leveraging on the above result, we propose to estimate DC functions by solving the following penalized least squares formulation:

$$\min_{\forall j, f_j \in \mathcal{DC}_j} \sum_{i=1}^{n} \left( y_i - \sum_{j=1}^{p} f_j(x_{ij}) \right)^2 + \sum_{j=1}^{p} (\lambda_d \|f_j\|_{\mathcal{DC}} + \lambda_s \|f_j\|_2), \qquad (11)$$

where the constant $\lambda_d$ controls the "degree of DC" of each $f_j$ while the constant $\lambda_s$ controls the number of effective components $f_j$ (i.e., sparsity). We see again here that the additive assumption is not only useful in interpretability but also in computability: it allows us to use the convenient complexity measure $\|\cdot\|_{\mathcal{DC}}$.

As before, we can simplify (11), which we call sparse difference of convex additive model (SD-CAM), as:

$$\min_{\forall j, \tilde{\mathbf{z}}_j \in \mathcal{H}} \sum_{i=1}^{n} \left( y_i - \sum_{j=1}^{p} z_{ij} \right)^2 + \sum_{j=1}^{p} \left( \lambda_d \|\tilde{\mathbf{z}}_j\|_{\mathcal{DC}_j} + \lambda_s \|\mathbf{z}_j\|_2 \right), \qquad (12)$$

where for $\mathbf{z} \in \mathbb{R}^n$ we define

$$\|\mathbf{z}\|_{\mathcal{DC}_j} := \sum_{i=2}^{n-1} \left| \frac{z_{i+1} - z_i}{\tilde{x}_{i+1,j} - \tilde{x}_{ij}} - \frac{z_i - z_{i-1}}{\tilde{x}_{ij} - \tilde{x}_{i-1,j}} \right|. \qquad (13)$$

Note that we have suppressed the dependence of $\|\cdot\|_{\mathcal{DC}_j}$ on $\tilde{\mathbf{x}}_j$ (the sorted version of $\mathbf{x}_j$). It is clear that we can easily incorporate further monotonic or convex constraints (or both) in (12), by adding

$\tilde{\mathbf{z}}_j \in \mathcal{I}$ (c.f. (4)) or $\tilde{\mathbf{z}}_j \in \mathcal{K}_j$ (c.f. (9)) or $\tilde{\mathbf{z}}_j \in \mathcal{K}'_j$ (c.f. (10)). On the other hand, the following variation may be useful in estimating *approximately* convex functions:

$$\|\mathbf{z}\|_{\mathcal{AC}_j} := \sum_{i=2}^{n-1} \max\left\{ \frac{z_i - z_{i-1}}{\tilde{x}_{ij} - \tilde{x}_{i-1,j}} - \frac{z_{i+1} - z_i}{\tilde{x}_{i+1,j} - \tilde{x}_{ij}}, 0 \right\}.$$

Indeed, $\|\mathbf{z}\|_{\mathcal{AC}_j} = 0$ iff the underlying function is convex. We note in passing that [Tib14] used a simplified form of (13) for trend filtering. However, the connection to DC functions and the implication for additive models are not investigated.

It is well-known in optimization that minimizing a DC function in general is intractable. However, our proposed procedure SDCAM for *estimating* a DC function is a tractable convex problem hence amenable to global and efficient algorithms, as we show next.

## 3.3 Backfitting Algorithm

We propose to solve SDCAM in (12) using a *modified* backfitting algorithm (i.e., block coordinate gradient). In each iteration, we fix all component fits except for one $\mathbf{z}_j$, and we solve the resulting subproblem:

$$\min_{\mathbf{z}_j \in \mathcal{H}} \tfrac{1}{2}\|\mathbf{r}_j - \mathbf{z}_j\|_2^2 + \lambda_d \|\tilde{\mathbf{z}}_j\|_{\mathcal{DC}_j} + \lambda_s \|\mathbf{z}_j\|_2, \tag{14}$$

where $\mathbf{r}_j \in \mathbb{R}^n$ is the partial residual that removes the contribution of $\mathbf{z}_j$. To show how we can solve (14), it is helpful to recall the following concept:

**Definition 1** (Proximal Map [Mor65]). *For any closed convex function $f : \mathbb{R}^n \to \mathbb{R} \cup \{\infty\}$, we define its proximal map as:*

$$\forall \mathbf{r} \in \mathbb{R}^n, \quad \mathsf{P}_f^\eta(\mathbf{r}) = \operatorname*{argmin}_{\mathbf{z} \in \mathbb{R}^n} \tfrac{1}{2\eta}\|\mathbf{z} - \mathbf{r}\|_2^2 + f(\mathbf{z}),$$

*and we write $\mathsf{P}_f = \mathsf{P}_f^1$.*

Note that the proximal map is a nonlinear map from $\mathbb{R}^n$ to $\mathbb{R}^n$. With this definition we can write the solution to (14) succinctly as $\mathsf{P}_{\lambda_d \|\cdot\|_{\mathcal{DC}_j} + \lambda_s \|\cdot\|_2 + \mathcal{H}}(\mathbf{r}_j)$, where we abuse the notation $\mathcal{H}$ as its indicator function

$$\iota_\mathcal{H}(\mathbf{z}) = \begin{cases} 0, & \text{if } \mathbf{z} \in \mathcal{H} \\ \infty, & \text{if } \mathbf{z} \notin \mathcal{H} \end{cases}.$$

Then, we can solve the subproblem (14) based on the following decomposition result:

**Theorem 4.** *The solution to subproblem (14) can be computed as*

$$\mathsf{P}_{\lambda_d \|\cdot\|_{\mathcal{DC}_j} + \lambda_s \|\cdot\|_2 + \mathcal{H}}(\mathbf{r}_j) = \mathsf{P}_{\|\cdot\|_2}^{\lambda_s}\left[\mathsf{P}_\mathcal{H}\big(\mathsf{P}_{\|\cdot\|_{\mathcal{DC}_j}}^{\lambda_d}(\mathbf{r}_j)\big)\right].$$

*Proof.* We first note that both $\lambda_d \|\cdot\|_{\mathcal{DC}_j}$ and $\mathcal{H}$ are positive homogeneous, hence we have

$$\mathsf{P}_{\lambda_d \|\cdot\|_{\mathcal{DC}_j} + \lambda_s \|\cdot\|_2 + \mathcal{H}}(\mathbf{r}_j) = \mathsf{P}_{\|\cdot\|_2}^{\lambda_s}\left[\mathsf{P}_{\lambda_d \|\cdot\|_{\mathcal{DC}_j} + \mathcal{H}}(\mathbf{r}_j)\right],$$

according to [Yu13, Theorem 4]. Next, by verifying the sufficient condition in [Yu13, Theorem 1], we have
$$\mathsf{P}_{\lambda_d \|\cdot\|_{\mathcal{DC}_j} + \mathcal{H}}(\mathbf{r}_j) = \mathsf{P}_{\mathcal{H}}\big(\mathsf{P}_{\|\cdot\|_{\mathcal{DC}_j}}^{\lambda_d}(\mathbf{r}_j)\big).$$
Combining the previous two claims we complete the proof. □

It is well-known that $\mathsf{P}_{\mathcal{H}}$ amounts to subtracting the average and $\mathsf{P}_{\|\cdot\|_2}^{\lambda_s}(\mathbf{r}) = \left(1 - \frac{\lambda_s}{\|\mathbf{r}\|_2}\right)_+ \mathbf{r}$, where $(t)_+ := \max\{t, 0\}$, is the block soft thresholding operator. Thus, we need only find a way to compute $\mathsf{P}_{\|\cdot\|_{\mathcal{DC}_j}}^{\lambda_d}(\mathbf{r}_j)$, i.e.,

$$\mathsf{P}_{\|\cdot\|_{\mathcal{DC}_j}}^{\lambda_d}(\mathbf{r}_j) = \underset{\mathbf{z} \in \mathbb{R}^n}{\operatorname{argmin}} \ \tfrac{1}{2}\|\mathbf{z} - \mathbf{r}_j\|_2^2 + \lambda_d \|\tilde{\mathbf{z}}\|_{\mathcal{DC}_j}.$$

With a suitable change of variables[4], the problem is equivalent to solving

$$\min_{s \in \mathbb{R}, \mathbf{w} \in \mathbb{R}^{n-1}} \ \frac{1}{2} \left\|A_j \begin{pmatrix} s \\ \mathbf{w} \end{pmatrix} - \tilde{\mathbf{r}}_j \right\|_2^2 + \lambda_d \|\mathbf{w}\|_{\mathsf{tv}}, \tag{15}$$

where

$$A_j = \begin{bmatrix} 1 & 0 & 0 & \cdots & 0 \\ 1 & \tilde{x}_{2j} - \tilde{x}_{1j} & 0 & \cdots & 0 \\ 1 & \tilde{x}_{2j} - \tilde{x}_{1j} & \tilde{x}_{3j} - \tilde{x}_{2j} & \cdots & 0 \\ \vdots & \vdots & \vdots & \ddots & \vdots \\ 1 & \tilde{x}_{2j} - \tilde{x}_{1j} & \tilde{x}_{3j} - \tilde{x}_{2j} & \cdots & \tilde{x}_{nj} - \tilde{x}_{n-1,j} \end{bmatrix}.$$

Since the proximal map $\mathsf{P}_{\lambda_d \|\cdot\|_{\mathsf{tv}}}^{\eta}$ can be computed in linear time using the algorithm in [DK01], we can solve (15) hence $\mathsf{P}_{\|\cdot\|_{\mathcal{DC}_j}}^{\lambda_d}(\mathbf{r}_j)$ iteratively using the accelerated proximal gradient algorithm [BT09, Nes13].

We summarize the entire computational procedure for SDCAM in Algorithm 1. The necessary modifications for the SCAM problem (8) should be clear: we only need to replace the proximal map $\mathsf{P}_{\lambda_d \|\cdot\|_{\mathsf{tv}}}^{\eta}$ by $\mathsf{P}_{\mathcal{I}}$ in line 10, i.e., projecting onto the isotonic cone $\mathcal{I}$ in (4). Importantly, note that the matrix $A_j$ appears only in two matrix-vector products $A_j^\top(\mathbf{z} - \mathbf{v})$ and $A_j\binom{s}{\mathbf{w}}$. By exploiting the special structure in $A_j$ we observe that we need *not* maintain $A_j$ explicitly (since there are only $n$ unique entries), and both matrix-vector products can be computed in $O(n)$ time using cumulative sums. Thus, the per-iteration complexity of Algorithm 1 is linear in the sample size $n$. Moreover, we can prove that Algorithm 1 converges to the global minimum at a sublinear rate by casting it under the block coordinate descent framework of [Tse01].

## 3.4 Interpolation

SDCAM, as well as any other nonparametric additive models, only yield the function value estimates $z_{ij} = f(x_{ij})$ on the observed training data $X$. As more and more data are observed, it is possible to estimate the entire function $f$ asymptotically. However, for a given finite sample size, there

---
[4]Define $s = \tilde{z}_1$ and $w_i = \frac{\tilde{z}_{i+1} - \tilde{z}_i}{\tilde{x}_{i+1,j} - \tilde{x}_{ij}}, i = 1, \ldots n-1$. Hence, $\tilde{\mathbf{z}} = A_j\binom{s}{\mathbf{w}}$ and $\|\tilde{\mathbf{z}}\|_{\mathcal{DC}_j} = \|\mathbf{w}\|_{\mathsf{tv}}$.

**Algorithm 1** Modified backfitting algorithm for Sparse Difference of Convex Additive Model (SDCAM)

**input:** $X \in \mathbb{R}^{n \times p}$, $\mathbf{y} \in \mathbb{R}^n$, $\lambda_d, \lambda_s \geq 0$, step size $\eta > 0$

1. $Z \leftarrow \mathbf{0}_{n \times p}$     // initialization
2. $\mathbf{r} \leftarrow \mathbf{y} - \mathsf{mean}(\mathbf{y})$     // remove mean $\mu$
3. **while** *not converged* **do**
4.     select some component $j \in \{1, 2, \ldots, p\}$
5.     $\mathbf{r} \leftarrow \mathbf{r} + Z_{\cdot j}$     // remove current component
6.     $\mathbf{v} \leftarrow \pi(\mathbf{r})$     // permute according to $\mathbf{x}_j$
7.     $\mathbf{z} \leftarrow \pi(Z_{\cdot j})$     // permute according to $\mathbf{x}_j$
8.     **for** $k = 1, \ldots, \max$ **do**
9.         $\binom{s}{\mathbf{w}} \leftarrow \binom{s}{\mathbf{w}} - \eta \cdot A_j^\top(\mathbf{z} - \mathbf{v})$     // grad step
10.        $\mathbf{w} \leftarrow \mathsf{P}^\eta_{\lambda_d \|\cdot\|_{\mathrm{tv}}}(\mathbf{w})$     // $\mathsf{P}_\mathcal{I}$ for SCAM (8)
11.        $\mathbf{z} \leftarrow A_j \binom{s}{\mathbf{w}}$     // substitution
12.     $\mathbf{z} \leftarrow \mathbf{z} - \mathsf{mean}(\mathbf{z})$     // computing $\mathsf{P}_\mathcal{H}$
13.     $\mathbf{z} \leftarrow \left(1 - \frac{\lambda_s}{\|\mathbf{z}\|_2}\right)_+ \mathbf{z}$     // computing $\mathsf{P}^{\lambda_s}_{\|\cdot\|_2}$
14.     $Z_{\cdot j} \leftarrow \pi^{-1}(\mathbf{z})$     // inverse permute
15.     $\mathbf{r} \leftarrow \mathbf{r} - Z_{\cdot j}$     // add component back

are infinitely many candidate function estimates that all yield the same fits $z_{ij}$ on the training data. To fix a particular choice, we can construct a function estimate by some form of smoothing or interpolation. In our experiments, we simply use linear interpolation to get a piece-wise linear function estimate, although in principle higher order interpolations can also be used. Note that we do not extrapolate the function values outside the range of the training data, because [BGS15] showed that a naive extrapolation can lead to large or even infinite generalization errors.

## 4 Experiments

In this section, we compare the performance of SDCAM to SpAM [RLLW08] and FLAM [PWS16] on both simulated and real data. SpAM is implemented using Gaussian kernel smoothers with the plug-in bandwidth. We do not compare to LISO [FM12] or SCAM because they achieve similar performance as SDCAM when the shape parameters are set correctly and perform much worse when the shape constraints do not apply hence are mis-specified.

### 4.1 Simulation Study

We generate samples $X \in \mathbb{R}^{n \times p}$ and $\mathbf{y} \in \mathbb{R}^n$ with $n = 100, p = 1000$ according to $y_i = \sum_{j=1}^4 f_j(x_{ij}) + \epsilon_i$, where $x_{ij} \sim \text{Uniform}(-2.5, 2.5)$ and $\epsilon_i \sim \mathcal{N}(0, \sigma^2)$. Two different values of $\sigma$ are chosen so that the signal-to-noise ratio SNR = 3 or 5, respectively. In other words, only the first 4 component functions are nonzero and the other 996 variables are irrelevant. We consider the following three scenarios for the choice of $f_1, \ldots, f_4$:

- Scenario 1: all of them are piecewise constant (Figure 1 (a)).

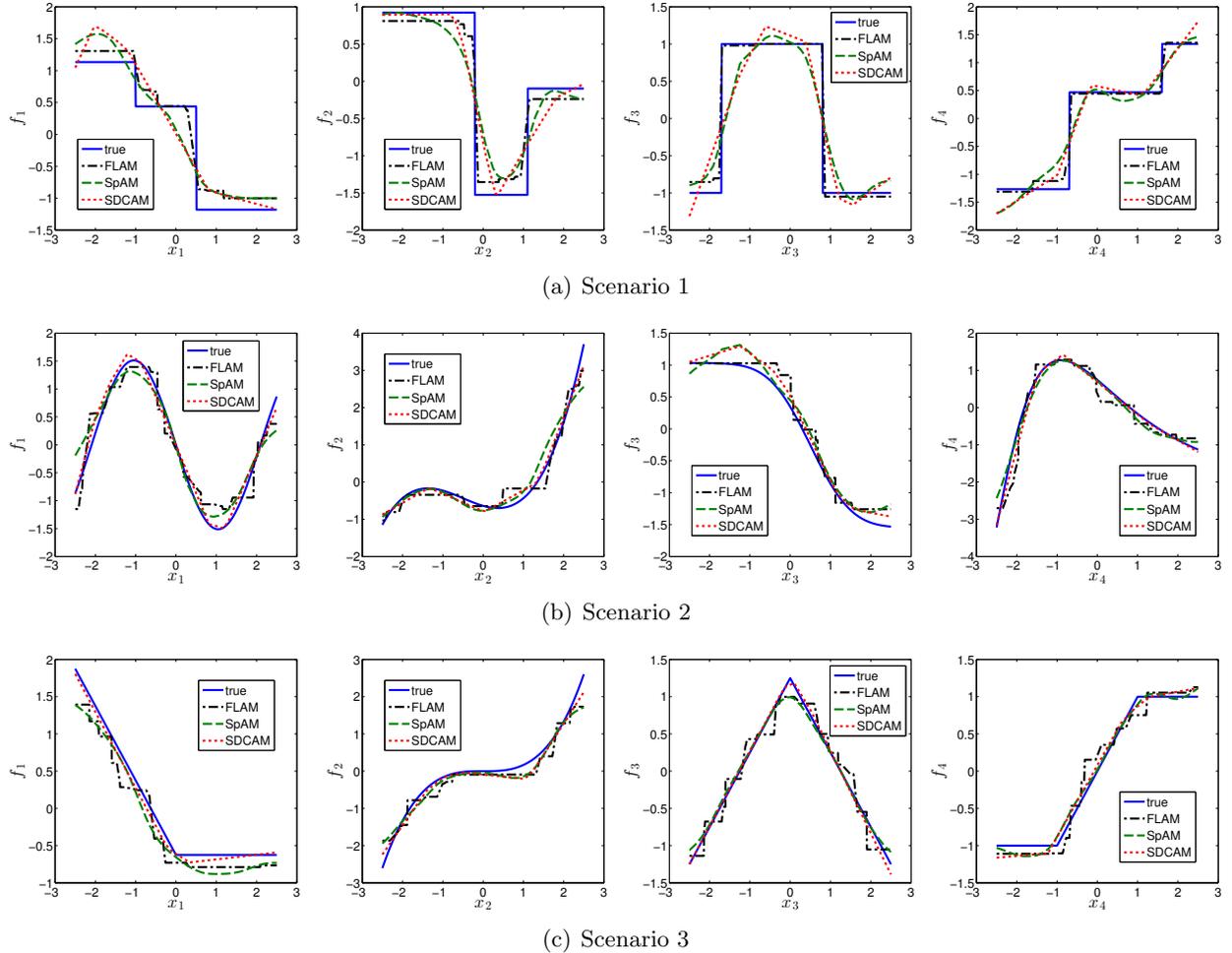

Figure 1: Typical examples of component fits $\hat{f}_1, \hat{f}_2, \hat{f}_3, \hat{f}_4$ in three simulation scenarios, with $n = 100$, $p = 1000$, and SNR $= 5$. The true component functions are plotted in solid blue lines. All models are estimated using the optimal tuning parameter (or parameter pair) that achieves the minimum validation set MSE.

- Scenario 2: all of them are smooth (Figure 1 (b)).

- Scenario 3: $f_2$ is smooth and the rest of them are piecewise linear (Figure 1 (c)).

For each combination of scenario and SNR level, we simulate training, validation, and test sets of the same size ($n = 100, p = 1000$). Validation sets are used to choose the optimal tuning parameter (or parameter pair) for all algorithms, and test sets are used to calculate the mean squared error (MSE) to evaluate the prediction accuracy of the estimated models. Table 1 summarizes the performance of each method in terms of precision/recall for sparsity pattern recovery, estimated model size, and test set MSE over 100 random data replicates.

As expected, FLAM achieves the lowest test set MSE in Scenario 1 because it employs a total variation regularizer to encourage the estimated functions to be piecewise constant (see also Figure 1 (a)). SDCAM performs comparably to SpAM in this scenario, in which neither is able to capture the sharp transition of values in the discontinuous functions.

| method | precision | recall | model size | MSE |
|---|---|---|---|---|
| Scenario 1, SNR = 3 | | | | |
| SpAM | 0.87 (0.19) | 0.99 (0.04) | 4.93 (1.86) | 1.15 (0.39) |
| FLAM | 0.82 (0.21) | 1.00 (0.00) | 5.48 (2.60) | 0.85 (0.24) |
| SDCAM | 0.86 (0.17) | 0.99 (0.05) | 4.78 (1.14) | 1.32 (0.41) |
| Scenario 1, SNR = 5 | | | | |
| SpAM | 0.92 (0.14) | 1.00 (0.00) | 4.54 (1.14) | 0.72 (0.20) |
| FLAM | 0.78 (0.22) | 1.00 (0.00) | 5.73 (2.27) | 0.43 (0.13) |
| SDCAM | 0.93 (0.13) | 1.00 (0.00) | 4.41 (0.96) | 0.75 (0.18) |
| Scenario 2, SNR = 3 | | | | |
| SpAM | 0.87 (0.19) | 0.99 (0.03) | 4.91 (1.52) | 0.89 (0.36) |
| FLAM | 0.80 (0.20) | 1.00 (0.00) | 5.48 (2.14) | 0.95 (0.29) |
| SDCAM | 0.93 (0.12) | 1.00 (0.00) | 4.38 (0.78) | 0.64 (0.14) |
| Scenario 2, SNR = 5 | | | | |
| SpAM | 0.92 (0.14) | 1.00 (0.00) | 4.51 (1.17) | 0.47 (0.17) |
| FLAM | 0.76 (0.19) | 1.00 (0.00) | 5.70 (1.74) | 0.49 (0.16) |
| SDCAM | 0.98 (0.06) | 1.00 (0.00) | 4.10 (0.30) | 0.25 (0.04) |
| Scenario 3, SNR = 3 | | | | |
| SpAM | 0.92 (0.16) | 1.00 (0.00) | 4.57 (1.39) | 0.47 (0.10) |
| FLAM | 0.86 (0.16) | 1.00 (0.00) | 4.84 (1.17) | 0.55 (0.11) |
| SDCAM | 0.97 (0.08) | 1.00 (0.00) | 4.15 (0.41) | 0.41 (0.06) |
| Scenario 3, SNR = 5 | | | | |
| SpAM | 0.98 (0.07) | 1.00 (0.00) | 4.11 (0.34) | 0.21 (0.04) |
| FLAM | 0.77 (0.19) | 1.00 (0.00) | 5.62 (1.85) | 0.26 (0.06) |
| SDCAM | 0.99 (0.05) | 1.00 (0.00) | 4.06 (0.24) | 0.15 (0.02) |

Table 1: Performance of different methods in the simulation study, with $n = 100$ and $p = 1000$. Shown are the mean (standard error) of precision/recall for sparsity pattern recovery, estimated model size, and test set MSE over 100 random data replicates.

In Scenario 2 where all relevant component functions are smooth, SDCAM outperforms SpAM, with FLAM being the worst option. Although SDCAM estimates seem to be quite similar to SpAM fits for the 4 true components (see Figure 1 (b)), SDCAM has the best precision score. That means, it is more effective in eliminating irrelevant variables and their spurious fits in the estimated models, which in turn can greatly improve the test set MSE.

Scenario 3 is particularly interesting: all the true component functions are DC but only $f_2$ is smooth. SDCAM achieves the best performance in such a mixed scenario, because SpAM cannot well handle the estimation of nonsmooth functions (in this case, piecewise linear functions) and smooth functions are unfavorable to FLAM. It is also worth noting that the SDCAM estimates tend to be less biased on the boundaries of the domain compared to SpAM fits, which are based on the smoothing technique (Figure 1 (c)).

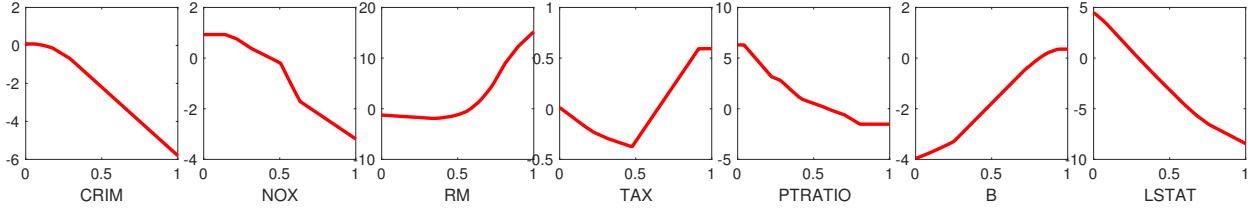

Figure 2: SDCAM estimates on the Boston housing data, with $p = 50$. Plotted are the fitted functions on the 7 selected variables. The remaining 43 components are all constant zero.

## 4.2 Real Data

We consider the Boston housing data that contains $n = 506$ observations of housing values (medv) in suburbs of Boston. Following [HSMW04] and [RLLW08], 10 original covariates are chosen to be included in the study: crim, indus, nox, rm, age, dis, tax, ptratio, b, lstat. We first rescale each variable to the interval $[0, 1]$, and then add spurious variables randomly drawn from Uniform$(0, 1)$ to produce three datasets of $p = 30, 40$, and $50$, respectively. For each such dataset, 10-fold cross-validation is used to select the optimal tuning parameter, and the final models are estimated on all the samples with the chosen value. Table 2 shows the proportion of 100 random cross-validation partitions that can correctly eliminate spurious variables in the final models. The performance of SDCAM is very consistent across different cross-validation partitions (and across different values of $p$) and is significantly better than the other two methods, demonstrating its superior ability in identifying irrelevant factors.

Because the SpAM and FLAM estimates are very sensitive to the choice of cross-validation partition, we only report the SDCAM fits for $p = 50$ in Figure 2. The results are largely compatible with two previous studies [RLLW08, FM12]: (1) the fitted patterns on selected crim, rm, ptratio, and lstat are quite similar; (2) indus, age, and dis are regarded to be irrelevant or have small effects; (3) there is a certain effect from nox, which was claimed to be large in [FM12] and borderline in [RLLW08]. Perhaps the most striking finding is a fairly large effect of tax, which were estimated to be insignificant in these two studies. Nevertheless, our function estimate of $f_7(\text{tax})$ is largely compatible with the fitted pattern reported in [HSMW04, Figure 8.3].

|        | SpAM | FLAM | SDCAM |
|--------|------|------|-------|
| $p = 30$ | 0.76 | 0.83 | 0.99  |
| $p = 40$ | 0.87 | 0.85 | 0.99  |
| $p = 50$ | 0.87 | 0.87 | 0.99  |

Table 2: The proportion of 100 random cross-validation partitions for the Boston housing data that can correctly eliminate irrelevant variables in the final models.

# 5  Conclusions

We have shown how to integrate shape constraints such as convexity into sparse additive models. The proposed sparse difference of convex additive model (SDCAM) can successfully estimate most additive continuous functions even in the presence of many irrelevant features. We propose a natural regularization functional in SDCAM to avoid overfitting and to reduce model complexity, and we develop an efficient backfitting algorithm with linear per-iteration complexity. Experiments on both synthetic and real data confirm that our method is competitive against state-of-the-art alternatives. Encouraged by our experiments, in the future we plan to study the model selection consistency of SDCAM and to investigate the bias-variance tradeoff when estimating a non-additive regression function.